\documentclass[lettersize,journal]{IEEEtran}
\usepackage{amsmath,amsfonts}
\usepackage{algorithm}
\usepackage{array}
\usepackage[caption=false,font=normalsize,labelfont=sf,textfont=sf]{subfig}
\usepackage{textcomp}
\usepackage{stfloats}
\usepackage{url}
\usepackage{verbatim}
\usepackage{graphicx}
\usepackage{cite}
\usepackage{multirow}
\newcommand{\cmmnt}[1]{}
\usepackage{booktabs}
\usepackage{amsmath}
\usepackage{algpseudocode}

\usepackage{amssymb}
\usepackage{bbding}
\usepackage{ragged2e}
\usepackage{threeparttable}
\usepackage{enumitem}
\usepackage{color}

\begin{document}

\title{PPD: A New Valet Parking Pedestrian Fisheye Dataset for Autonomous Driving}

\author{Zizhang Wu, Xinyuan Chen, Fan Song, Yuanzhu Gan, Tianhao Xu, Jian Pu*, Rui Tang*
\thanks{This paper was produced by Fudan University and Zongmu Technology. Zizhang Wu, Xinyuan Chen and Fan Song contributed equally to this work. Corresponding author: Jian Pu, Rui Tang.

Zizhang Wu, Jian Pu (e-mail:wuzizhang87@gmail.com; jianpu@fudan.edu.cn) is with Fudan University, No. 220, Handan Road, Shanghai, China.

Xinyuan Chen, Fan Song, Yuanzhu Gan, Tianhao Xu and Rui Tang are with the Zongmu Technology (Shanghai) Co., Ltd, Building 10, Zhangjiang Artificial Intelligence Island, Lane 55, Chuanhe Road, Pudong New Area, Shanghai, China. (e-mails: lan.chen@zongmutech.com; fan.song@zongmutech.com; yuanzhu.gan@zongmutech.com; tobias.xu@zongmutech.com; rui.tang@zongmutech.com)

}
\thanks{Manuscript received xx xx, 2023; revised xx xx, 2023.}}

\markboth{Journal of \LaTeX\ Class Files,~Vol.~14, No.~8, August~2023}%
{Shell \MakeLowercase{\textit{et al.}}: A Sample Article Using IEEEtran.cls for IEEE Journals}


\maketitle

\begin{abstract}
Pedestrian detection under valet parking scenarios is fundamental for autonomous driving. 
However, the presence of pedestrians can be manifested in a variety of ways and postures under imperfect ambient conditions, which can adversely affect detection performance. 
Furthermore, models trained on public datasets that include pedestrians generally provide suboptimal outcomes for these valet parking scenarios. 
In this paper, we present the \textbf{P}arking \textbf{P}edestrian \textbf{D}ataset (\textbf{PPD}), a large-scale fisheye dataset to support research dealing with real-world pedestrians, especially with occlusions and diverse postures.
\textbf{PPD} consists of several distinctive types of pedestrians captured with fisheye cameras.
Additionally, we present a pedestrian detection baseline on \textbf{PPD} dataset, and introduce two data augmentation techniques to improve the baseline by enhancing the diversity of the original dataset. 
Extensive experiments validate the effectiveness of our novel data augmentation approaches over baselines and the dataset's exceptional generalizability.
\end{abstract}


\section{Introduction}
To develop an advanced driver assistance system (ADAS) that is both effective and safe for parking lot scenarios~\cite{wu2021deepword, wu2020psdet,tiv1,tiv3,tiv4}, 
it is critical to ensure the safety of road users such as pedestrians.
%
The detection range of conventional pinhole cameras is often insufficient to detect the variety of behaviors and postures displayed by pedestrians.
As an alternative to pinhole cameras, fisheye cameras could have a wider field of vision (FoV)~\cite{2018IVreal}, which is necessary for the perception of close range and low altitude, particularly in a traffic bottleneck.
Thus, fisheye cameras are becoming increasingly prominent in driverless vehicles as intelligent alternatives to traditional cameras.




Nevertheless, such pedestrian detection \cite{zhao2019object, article41,article42,tiv6} still remains difficult due to evasive irregular postures and imprecise surrounding circumstances.
First, there is a wide range of pedestrian behaviors that are rarely represented in publicly available datasets, such as occlusion, lying down, walking, etc.
%
Second, the fisheye lens's radial distortion leads to substantial appearance distortion \cite{wu2021disentangling,tiv5}, complicating the pedestrian recognition process.
Additionally, the quality of images is significantly affected by environmental factors such as light and opacity. 

\begin{figure}[!t]
\centering
\includegraphics[width=0.42\textwidth]{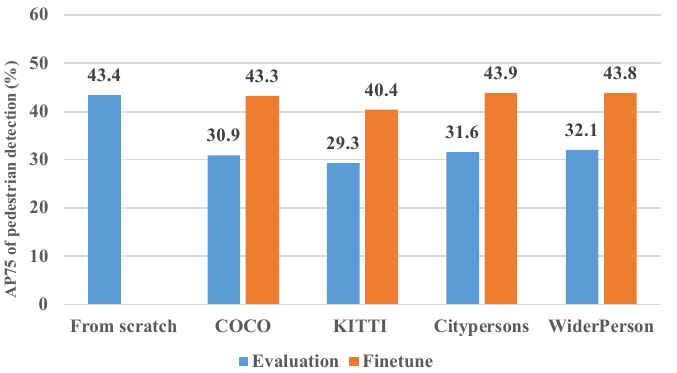}
\caption{
The cross-dataset testing on the \textbf{PPD} dataset.
``Evaluation" refers to explicitly evaluating public datasets' pre-trained models on the \textbf{PPD} dataset, with suboptimal results compared with the model trained from scratch.
``Finetune" means finetuning these pre-trained models on the \textbf{PPD} dataset with little advancement.}
\label{fig:first}
\end{figure}

\begin{figure*}
  \includegraphics[width=0.65\textwidth]{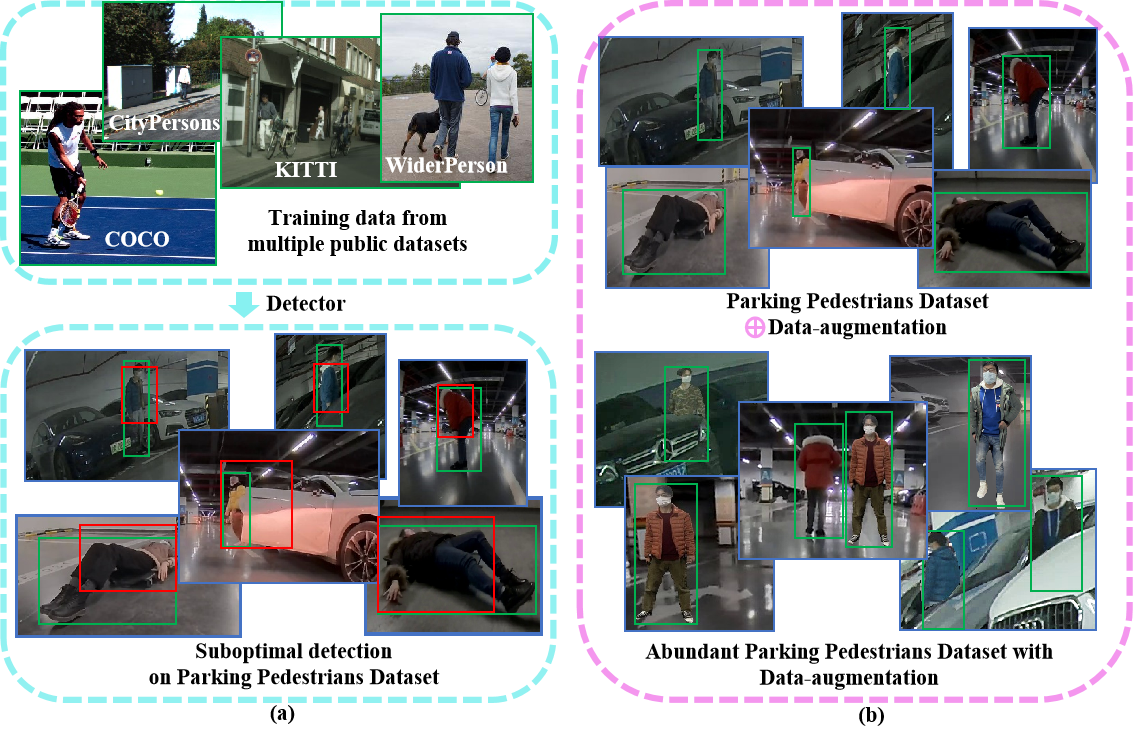}
  \centering
  \caption{
  Green boxes indicate ground truth and red boxes indicate predictions.
  (a) The suboptimal performance of the detector pre-trained on public datasets in parking lot scenes. (b) Our \textbf{P}arking \textbf{P}edestrian  \textbf{D}ataset (PPD) uses data augmentation methods and provides diverse large real-world pedestrian data with occlusion and different postures.}
  \label{fig:teaser}
\end{figure*}

Current datasets and benchmarks for pedestrian detection, including Caltech USA \cite{article21}, KITTI \cite{article22}, CityPersons \cite{article24}, and Wider-Person \cite{article23}, have aided in rapid progress for the pedestrian detection task. 
These datasets usually encompass urban, rural, or highway driving scenes, and their pinhole cameras comfortably capture high-grade images with clear and distinguishable pedestrians.  
Moreover, Valeo delivers the fisheye automotive dataset WoodScape with extensive content \cite{yogamani2019woodscape}.
However, public datasets place insufficient emphasis on pedestrians with irregular postures and fisheye image formation.
The models trained on public datasets reveal suboptimal performance in difficult parking scenes without a large number of training instances, as shown in Fig.\ref{fig:first} and Fig.\ref{fig:teaser}. 

To expand real-world fisheye pedestrians' images with various occlusion and postures under valet parking scenes, this paper offers a new large-scale fisheye dataset called \textbf{P}arking \textbf{P}edestrian \textbf{D}ataset (\textbf{PPD}) to promote the research on pedestrian problems, as shown in Figure \ref{fig:teaser} (b). 
Different from other pedestrian datasets \cite{article21,article22,article23,article24}, our \textbf{PPD} dataset focuses on pedestrian detection and provides more than 330K fisheye images in valet parking scenes.
To guarantee the pedestrians' diversity, we collect data from different parking lots, various periods, and diverse pedestrian situations.
Additionally, we subdivide \textbf{PPD} into three sub-datasets: Occlusion Pedestrians (\textbf{OP}), Posture Pedestrians (\textbf{PP}), and Lying-down Pedestrians (\textbf{LP}). 
\textbf{OP} involves pedestrians occluded by car bodies, doors, or parking lots' pillars. 
\textbf{PP} is concerned with pedestrians' abundant postures, including standing, stooping, sitting, etc.
\textbf{LP} concentrates on lying-down pedestrians, the most perilous situation that requires immediate early warning.

To reduce annotation costs and further broaden the diversity of pedestrians, we further propose two data augmentation techniques:
%
\textbf{O}cc-\textbf{D}ata-\textbf{A}ugmentation (\textbf{ODA}) and \textbf{P}os-\textbf{D}ata-\textbf{A}ugmentation (\textbf{PDA}) for pedestrians' occlusions and postures, respectively.
Using \textbf{ODA} and \textbf{PDA}, high-quality synthetic images are generated through the collecting, localization, and local fusion procedures, complementing the commonly used hybrid augmentation methods~\cite{cherian2021image, liesting2021data, article50}. 
%
%
Besides, we build pedestrian detection baselines on our \textbf{PPD} dataset and extensive experiments validate the effectiveness of our novel data augmentation approaches. In addition, the cross-dataset evaluation reveals \textbf{PPD}'s exceptional capacity to generalize.

Our contributions are summarized as follows:
\begin{itemize}
\item We provide the first fisheye dataset comprising over 330K fisheye images, particularly for diverse occlusion and postures of pedestrians in parking lot scenes. 
\item We report the baseline results of pedestrian detection on the proposed \textbf{PPD} dataset and propose two novel data augmentation techniques to improve the baseline.
\item  Extensive experiments demonstrate the effectiveness of \textbf{ODA}, \textbf{PDA}, and \textbf{PPD}'s exceptional generalizability.
\end{itemize}
\section{Related Work}
In this section, we briefly introduce the related works to our topic, i.e., pedestrian detection dataset, pedestrian detection frameworks, and data augmentation methods. 
\begin{table*}[t]
\caption{Statistics of our collected parking pedestrian dataset, which consists of three subdatasets: Occlusion Pedestrian Dataset, Posture Pedestrian Dataset and Lying-down Pedestrian Dataset. ``\#" denotes the number.
}
\label{tab:dataset_info}
\centering
\resizebox{\linewidth}{!}{
\begin{tabular}{c|c|c|cccccc}
\toprule
\multirow{2}{*}{Dataset}             & \multirow{2}{*}{Sub-dataset} & \multirow{2}{*}{Image resolution} & \multicolumn{3}{c}{\#images} & \multicolumn{3}{c}{\#pedestrians} \\ \cline{4-9} 
                                     &                               &                       & train     & val      & test     & train    & val     & test    \\ \midrule
\multirow{3}{*}{Parking Pedestrians} & Occlusion Pedestrians (\textbf{OP})         & 1280×580              & 51921     & 31900     & 27700     & 208659    & 105303    & 104211    \\
                                     & Posture Pedstrians (\textbf{PP})            & 1920×870              & 52123      & 39080      & 27021        & 206401    & 140800     & 123120       \\
                                     & Lying-down Pedestrians (\textbf{LP})        & 1920×870              & 53993       & 39022      & 22921       & 205300      & 103978     & 90090       \\ \bottomrule
\end{tabular}}
\end{table*}
\subsection{Pedestrian Detection Datasets}
Pioneer works of pedestrian detection datasets involve \cite{lin2014microsoft,article21,article22,article23,article24,dollar2009pedestrian,hwang2015multispectral,ouyang2012discriminative,wu2005detection}, which contribute to great progress in pedestrian detection. 
There are large-scale datasets such as Caltech USA \cite{article21} and KITTI \cite{article22}, which contain urban, rural, and highway scenes and provide annotation frame sequences on videos. 
However, both datasets have low pedestrian densities.
More recently, researchers proposed vast and diversified datasets, WiderPerson \cite{article23} and CityPersons \cite{article24}. 
CityPersons \cite{article24} is the subset of the CityScapes Dataset, whose average pedestrian density grows three times that of KITTI \cite{article22}. 
WiderPerson \cite{article23} contains a range of scenarios, including marathon, traffic, activity, dance, and miscellany, totaling approximately 400 thousand annotated instances.
Moreover, Valeo proposed the extensive automotive dataset WoodScape \cite{yogamani2019woodscape} with fisheye cameras instead of pinhole cameras.
However, there is no publicly available benchmark dataset for valet parking pedestrian scenarios,
particularly those including varied occlusions and postures, where suboptimal detection of pedestrians forms a threat to driving safety.
\subsection{Pedestrian Detection Frameworks} 
CNN-based pedestrian detection methods can be generally categorized into one-stage~\cite{article3,article4,article33} and two-stage \cite{article10,article34} methods. 
As an end-to-end pipeline, one-stage methods achieve a significant trade-off between performance and speed, such as the SSD series \cite{article28,liu2016ssd,article45,article49}, YOLO series \cite{article4,article5,article33,tiv2} and Retinanet \cite{article3}.
In contrast, two-stage methods, such as the RCNN series \cite{article7,article10,he2017mask,article34}, take advantage of the predefined anchors to improve the performance at the cost of speed. 
Furthermore, recent works~\cite{article31,article44,article46,article47} fuse multi-scale feature maps to improve pedestrian detection with different scales. 
Moreover, other works~\cite{article19,article20,article41,article42,zhang2018occluded} focus on crowded pedestrian detection problems for actual applications.
\cite{article19}
designed a new boundary box regression loss specifically for better pedestrian localization. 
Liu et al. offer a non-maximum suppression method to refine the bounding boxes given by detectors\cite{article20}.

\subsection{Data Augmentation Methods}
Data Augmentation methods such as random cropping, color dithering and flipping, play an important role in achieving state-of-the-arts \cite{devries2017improved,shorten2019survey,arazo2020pseudo}. These enhancements are more generic in nature and are particularly relevant to expressing data transformation invariance.
In addition, hybrid image augmentation \cite{cherian2021image, liesting2021data, article50} can mix cross-image information, which usually applies appropriate changes with labels to increase diversity. 
Furthermore, the adaptations of mixups \cite{yun2019cutmix, article50,dwibedi2017cut} are popular among hybrid image augmentation. 
CutMix \cite{yun2019cutmix} pastes a rectangular crop of the image instead of mixing all pixels. 
It creates new composite pictures by combining the rectangular grids of individual images with actual boxes.
Cut-Paste-and-Learn \cite{dwibedi2017cut} 
extracts objects in poses and then mixes and pastes them to different backgrounds. 
Copy-paste \cite{article50} fuses information from different images in an object-aware manner: copying and pasting instances across images. 

\section{Parking Pedestrians Dataset}
In this section, we introduce our \textbf{P}arking \textbf{P}edestrian \textbf{D}ataset (\textbf{PPD}) dataset in detail, including the data collection, annotation protocols, informative statistics, and dataset characteristics.
\subsection{Data Collection}

To ensure the diversity of pedestrians, we collect data from 3 cities, 12 parking lots, two periods (morning and evening),
 and different pedestrians with various ages, heights, clothes, and postures.
In total, we captured 100 videos that last from 1 hour to 6 hours and with an average of 2 hours.
%
Then, we convert the videos into pictures and select the images containing pedestrian instances.
For high-quality images, we restrict the visible range of the fisheye camera and further remove distorted and blurred pedestrian images.
Also, we do not cover all pedestrians' continuous moving processes for redundant annotations.
Instead, we select the best-quality images and then apply our data augmentation methods (discussed later) to increase the data variance.
Based on the images' content, we also divide them into three categories: occlusion pedestrians, posture pedestrians, and lying-down pedestrians.

Table \ref{tab:dataset_info} illustrates the statistics of the \textbf{PPD} dataset. 
A total of more than 330K images comprise three sub-datasets: Occlusion Pedestrians Dataset, Posture Pedestrians Dataset and Lying-down Pedestrians Dataset, with amounts of 111,521, 118,224 and 115,936, respectively. 
Besides, every sub-dataset further performs partitioning into training, validation and testing sets at a ratio of 5:3:2. 

\subsection{Image Annotation}
We annotate the dataset in the same way as the Caltech dataset \cite{dollar2009pedestrian}, by drawing a tight bounding box around each pedestrian's complete body.
However, occluded pedestrians are special since the foreground-like car bodies or parking lot pillars often lead to incomplete pedestrian instances.
Therefore, we have to estimate the distance from the pedestrian instance to the car's fisheye camera, and then roughly calculate the size of the box according to the depth proportion, as shown below:
\begin{equation}
\label{equ1}
    W_{o} = W_{p} \times (1 - D_o / D_{max}),
\end{equation}
\begin{equation}
\label{equ2}
    H_o = H_p \times (1 - D_o / D_{max}),
\end{equation}
where $H_p$ and $W_p$ are the average human height and width, predefined as 1.7 meters and 0.3 meters, respectively. 
$D_o$ is the depth from the occluded pedestrian instance to the camera.
Since the parking space has a fixed size, we can estimate the depth approximately by the relative location between the pedestrian instance and the nearby parking space within the same image. 
$D_{max}$ is the max depth of the fisheye camera. 
Finally, based on the depth ratio between $D_o$ and $D_{max}$, we can roughly infer the annotated width $W_o$ and height $H_o$, as shown in Equations. (\ref{equ1}) and (\ref{equ2}).

\begin{figure*}[!h]
\centering
\includegraphics[width=0.91\textwidth]{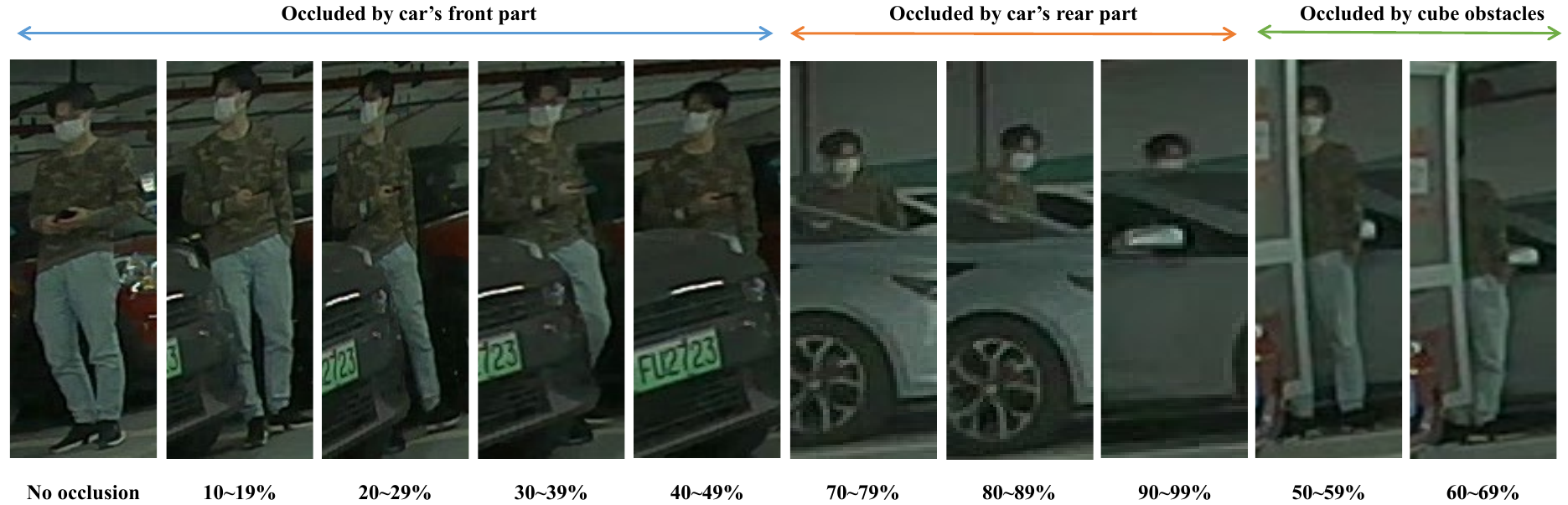}
\caption{Examples of different occlusion scales and occlusion types in the Occlusion Pedestrian Dataset.}
\label{fig:occradio}
\end{figure*}

\begin{figure}[!t]
\includegraphics[width=0.48\textwidth]{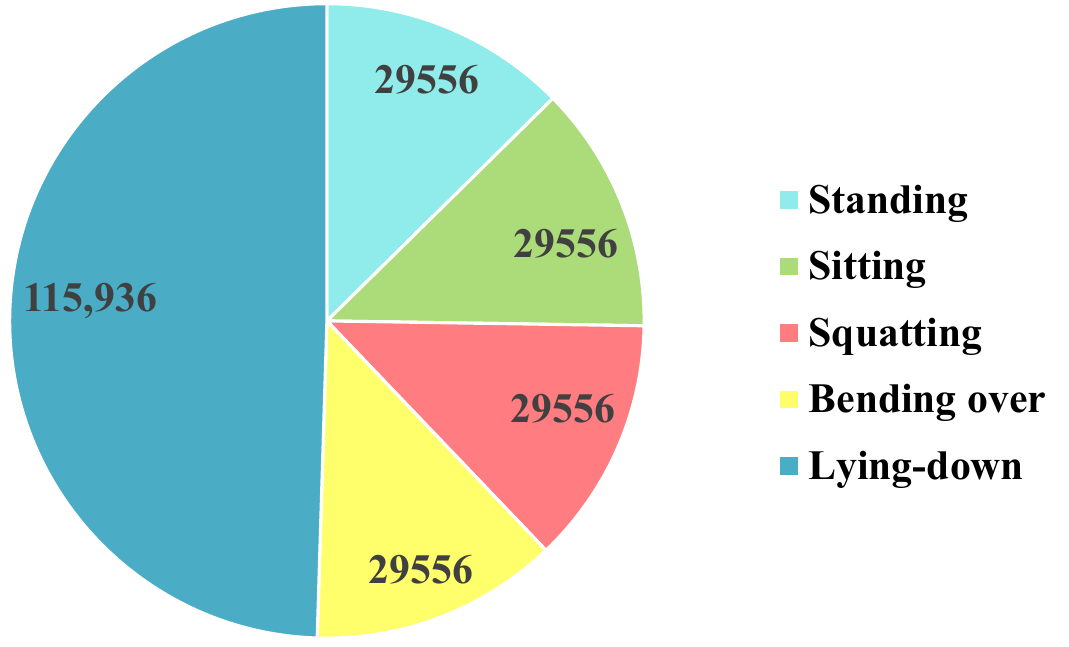}
\caption{The detailed distribution of posture pedestrian dataset and lying-down pedestrians dataset.}
\label{fig:datavolume} 
\end{figure}

\subsection{Sub-datasets Description}
The occluded pedestrian dataset provides three occlusion scenarios with different occlusion rates.
As shown in Fig. \ref{fig:occradio}, to better restore reality, we collect the pedestrians occluded by the cars' part and cube obstacles with 10 occlusion rates, starting from not occluded to 99\% with 10\% increments per class.

The posture pedestrian dataset contains four postures: standing, sitting, squatting and bending over. We strive to cover eight pedestrians' orientations, which are front, rear, left, right, left front, right front, left rear, and right rear. 
In addition, we divide lying-down pedestrians into the new subset since they are the most dangerous cases. We make an effort to cover the same eight orientations as the posture subset.
The detailed distribution of posture and lying-down pedestrians is shown in Fig. \ref{fig:datavolume}, and a detailed explanation of \textbf{PPD}'s categories is reported in Table \ref{tab:category_ppd}.

Furthermore, to further broaden pedestrians' diversity, we apply two novel data augmentation techniques to occlusion and posture pedestrians. 
The data volume increases, and later experiments will indicate improvements in pedestrian detection performance.
\begin{table}[!t]
\caption{The explanations for \textbf{PPD}'s categories.}
\label{tab:category_ppd}
\centering
\resizebox{\linewidth}{!}{
\begin{tabular}{ccc}
\toprule
\multicolumn{2}{c|}{\textbf{Category}}                                               & \textbf{Explanation}                                   \\ \midrule
\multicolumn{1}{c|}{\multirow{3}{*}{Occlusion}} & \multicolumn{1}{c|}{Car's front part}        & Pedestrians occluded by car's front part                       \\ \cline{2-3} 
\multicolumn{1}{c|}{}                           & \multicolumn{1}{c|}{Car's rear part}       & Pedestrians occluded by car's rear part                         \\ \cline{2-3} 
\multicolumn{1}{c|}{}                           & \multicolumn{1}{c|}{Cube obstacle}         & Pedestrians occluded by cube obstacles like pillar or fire hose                      \\ \midrule
\multicolumn{1}{c|}{\multirow{4}{*}{Posture}}   & \multicolumn{1}{c|}{Standing}     & Pedestrians with standing posture                       \\ \cline{2-3} 
\multicolumn{1}{c|}{}                           & \multicolumn{1}{c|}{Sitting}      & Pedestrians with sitting posture                        \\ \cline{2-3} 
\multicolumn{1}{c|}{}                           & \multicolumn{1}{c|}{Squatting}    & Pedestrians with squatting posture                      \\ \cline{2-3} 
\multicolumn{1}{c|}{}                           & \multicolumn{1}{c|}{Bending over} & Pedestrians with bending over posture                   \\ \midrule
\multicolumn{2}{c|}{Lying-down}                                                     & Lying-down pedestrians, independent of posture pedestrians \\ \bottomrule
\end{tabular}}
\end{table}

\subsection{Dataset Characteristics}
Our \textbf{PPD} dataset exhibits differences from public datasets in image representation style, scenarios, quantity and diversity.
Below, we elaborate on the four main characteristics of our \textbf{PPD} dataset.
\\

\noindent{\textbf{Fisheye image representation.}}
The \textbf{PPD} dataset consists of fisheye images, different from the common pinhole images of public datasets. 
Fisheye images provide a larger field-of-view (FoV), which is more suitable for close-range and low-lying pedestrian detection.\\

\noindent{\textbf{Specific parking scenarios.}}
The \textbf{PPD} dataset focuses on pedestrian detection in parking scenarios, which is also distinct from natural scenes of public datasets.
The environmental conditions in parking scenarios, such as light and opacity, significantly increase the detection difficulty.
Concerning a variety of tough pedestrian scenarios, \textbf{PPD} can promote research in dealing with real-world pedestrian problems.\\

\noindent{\textbf{Large quantity.}}
Our \textbf{PPD} dataset obtains more than 330 thousand data samples from more than 200-hour parking scene video clips. 
We constantly collect diverse parking pedestrian scenarios, eventually reaching the goal of over one million data.\\

\noindent{\textbf{High quality and diversity.}}
Our \textbf{PPD} dataset covers 3 cities, 12 parking lots from different periods and different pedestrian cases. Additionally, we carefully select high-quality images with high resolution and apply data augmentation techniques to enlarge diversity.

\section{The Proposed Data Augmentations}
Training detectors for special pedestrians usually requires a large quantity of data, which demands tremendous resources and effort to acquire and annotate.
Considering these challenges, we provide two novel data augmentation techniques: \textbf{O}cc-\textbf{D}ata-\textbf{A}ugmentation (\textbf{\textbf{ODA}}) and \textbf{P}os-\textbf{D}ata-\textbf{A}ugmentation (\textbf{PDA}).
Specifically, \textbf{\textbf{ODA}} focuses on occluded pedestrians, and \textbf{PDA} targets pedestrians with different postures.
In this section, we describe those two data augmentation methods in detail.


\begin{figure*}[t]
\centering
\includegraphics[width=0.82\textwidth]{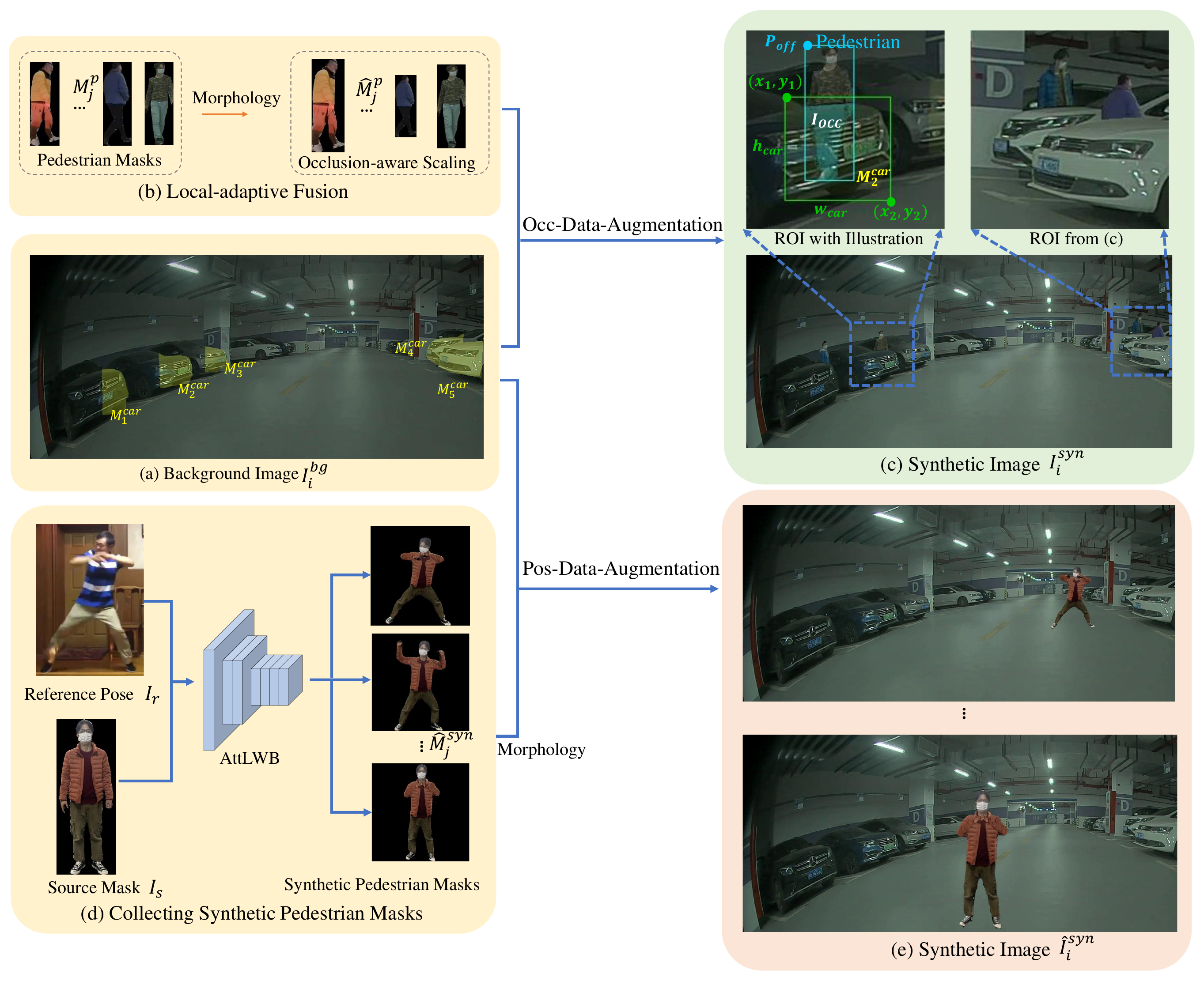}  
\caption{Overview of our data augmentation methods. 
Our methods include \textbf{O}cc-\textbf{D}ata-\textbf{A}ugmentation (\textbf{\textbf{ODA}}) and \textbf{P}os-\textbf{D}ata-\textbf{A}ugmentation (\textbf{PDA}).
\textbf{ODA} and \textbf{PDA} have the same pipeline: (1) collecting pedestrian masks and background images; (2) determining where to paste pedestrian masks; (3) fusing pedestrian masks with background images.}
\vspace{-1mm}
\label{fig:overview}    
\end{figure*}

\subsection{Overall Pipeline}
We define our data augmentation process as $f(*)$, so the overall structure states are as follows:
\begin{equation}
    I^{syn}_i = f(I^{bg}_i, M_j), i=1,2,\cdots,N, j=1,2,\cdots,K
\label{eq:total}
\end{equation}
where $I^{bg}_i$ is the background image, $M_j$ indicates pedestrian masks and $I^{syn}_i$ indicates the produced synthetic images. 

As shown in Fig. \ref{fig:overview}, our augmentation pipeline contains three stages: (1) collecting pedestrian masks and background images; (2) determining where to paste pedestrian masks; and (3) fusing pedestrian masks with background images.

\subsection{Occ-Data Augmentation}
We present the \textbf{O}cc-\textbf{D}ata-\textbf{A}ugmentation (\textbf{\textbf{ODA}}) method for occluded pedestrians with three procedures. The detailed procedure can be found in Algorithm 1. 

\begin{algorithm}[!h]
\caption{Occ-Data-Augmentation}
\label{alg:Occ-Pseudo-labeling}
\begin{algorithmic}[1]
\Require pedestrian masks $M^{p}_{j},j=1,2,\cdots,K$; background images $I^{bg}_{i},i=1,2,\cdots,N$; front car parts' masks $M^{car}_{i}, i=1,2,\cdots,N$
\Ensure $I^{syn}_i, i=1,2,\cdots,N$
\State Fix random seed;
\State Shuffle $I^{bg}$ and $M^{car}$;
\State $i \gets 1$;
\While{$i <= N$}
    \State $\widehat{M}^{car}_t, t=1,2,\cdots,T \gets$ randomly select some of labels from $M^{car}_{i}$;
    \For{$\widehat{M}^{car}_t$ in $\widehat{M}^{car}$}
        \State $x_1,y_1,x_2,y_2 \gets \widehat{M}^{car}_t$'s shape;
        \State $w_{car},h_{car} = x_2-x_1,y_2-y_1$;
        \State $M^{p}_j \in \mathbb{R}^{w_p \times h_p} \gets$ randomly select from $M^{p}$;
        \State Resize $M^{p}_j$ to $\widehat{M}^{p}_j$ with $w^{\prime}_{p}=w_{p}\frac{h_{car}}{h_{p}}$, $h^{\prime}_{p}=h_{car}$; 
        \State $P_{off}=(randint(x_1, x_2 - w^{\prime}_p), y_1 - randint(0.2*h_{car}, 0.3*h_{car}))$;
        \State $I_{occ} = \widehat{M}^{p}_j * \widehat{M}^{car}_t$ in $P_{off}$;
        \State $I^{fg}_{avil} = \widehat{M}^{p}_j - I_{occ}$;
        \State $I^{syn}_i = \alpha * I^{fg}_{avil} + (1 - \alpha) * I^{bg}_i$;
        \State Create pseudo label of $I_{syn}$;
    \EndFor
\State $i \gets i + 1$;
\EndWhile
\State \Return $I^{syn}$.
\end{algorithmic}
\end{algorithm}

\subsubsection{Collecting pedestrian masks and background images}
We observe that in the valet parking scenes, cars' front or rear parts mostly occlude pedestrians. To address such occlusion, our \textbf{ODA} requires the masks of pedestrians $M^{p}_{j},j=1,2,\cdots,K$ as foreground and the background images $I^{bg}_{i},i=1,2,\cdots,N$, containing the masks of cars' front parts $M^{car}_{i}, i=1,2,\cdots,N$. 

For a background image, we label all available cars' front parts with the Labelme tool \cite{russell2008labelme}, as shown in Fig. \ref{fig:overview} (a). Note that the pedestrian masks should be diverse and high quality, which determines the reality of the synthesized images. Thus, we process them with morphological operations such as OPEN and ERODE. To seamlessly paste pedestrian masks into background images, we apply occlusion-aware scaling, resizing the mask to a more realistic scale.


\subsubsection{Localization of synthetic occlusion}
Following Algorithm 1, we take one background image $I^{bg}_{i}$ as input and randomly pick one car's front part $\widehat{M}^{car}_t$ as the pasting location.
Then, we prepare the pedestrian mask $M^{p}_j$, randomly selected from the mask list $M^{p}$.

\subsubsection{Local fusion for occlusion}
\begin{figure}[!t]
\centering
\includegraphics[width=0.65\linewidth]{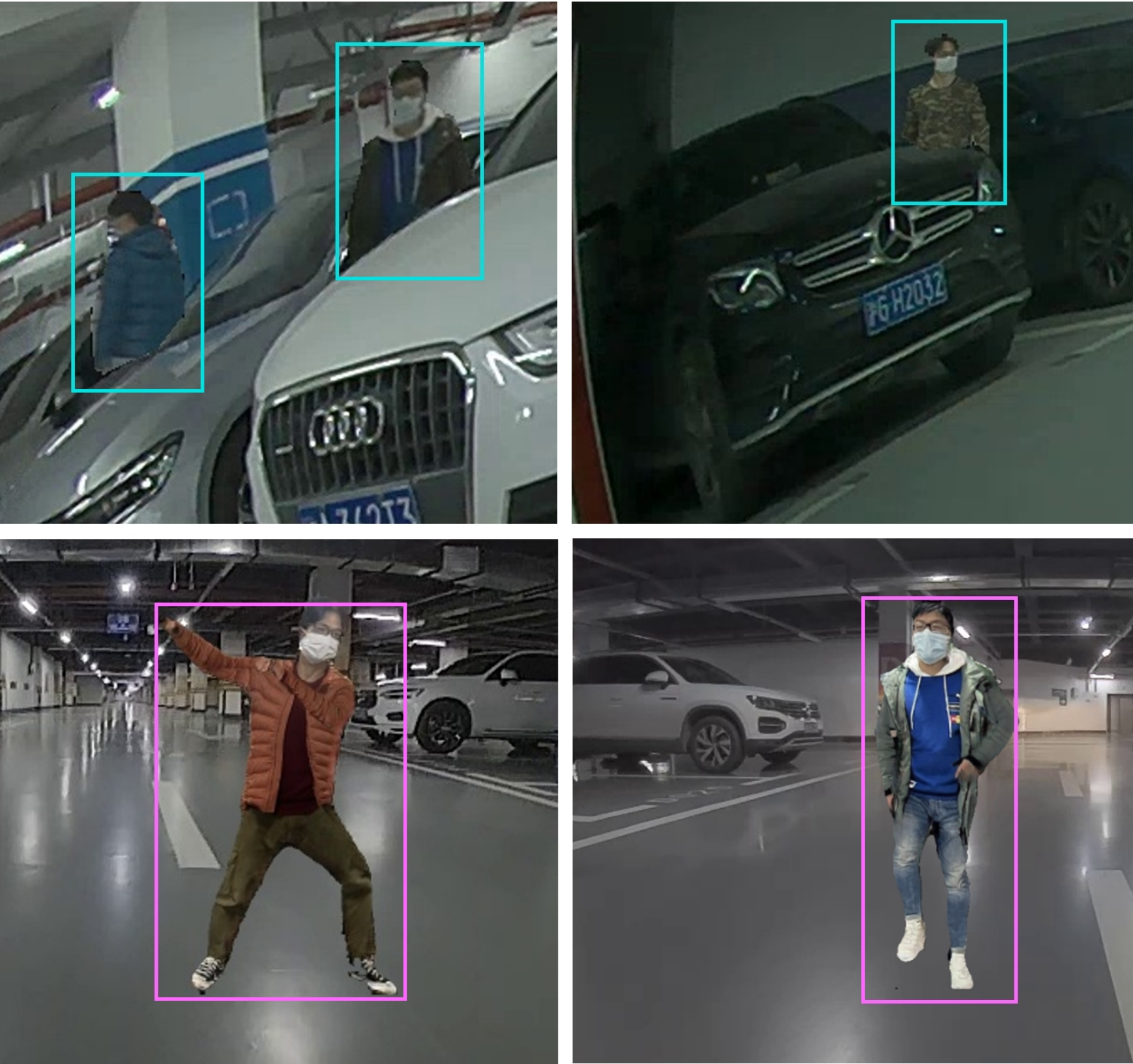}
\caption{Examples of augmented images. Top row: ODA results; Bottom row: PDA results.}
\label{fig:example_PL}   
\vspace{-2mm}
\end{figure}

For more precise localization, we paste the top-left point $P_{off}$ of the pedestrian mask above the car's front part $\widehat{M}^{car}_t$ with random but limited distances, as shown in Fig. \ref{fig:overview}(d). 
Specifically, we calculate $P_{off}$ according to:
\begin{equation}
    P_{off} = \left [ \begin{matrix}
    x \\
    y \\
    \end{matrix} \right ]  =
    \left [ \begin{matrix}
    randint(x_1, x_2 - w^{\prime}_p) \\
    y_1 - randint(0.2*h_{car}, 0.3*h_{car})\\
    \end{matrix} \right ]
\label{eq:locationP}
\end{equation}
where $w^{\prime}_p$ is the width of the resized pedestrian mask and $(x_1, y_1)$, $(x_2, y_2)$ are the top-left and bottom-right points of the front car part's mask. 
$h_{car}$ and $w_{car}$ are the corresponding width and height of the front car part's mask, respectively.
It is noteworthy that all coordinates are relative to the top-left vertex. 
Furthermore, we remove the intersection between the pedestrian mask and the background, that is, the occlusion region: $I_{occ} = \widehat{M}^{p}_j \cap \widehat{M}^{car}_t$ (line 13 of Algorithm 1).
In this way, we accomplish the pseudo occluded pedestrian's mask $ I^{fg}_{avil}$.
Then, we paste the pedestrian's mask into the background image and update the label, according to
\begin{align}
I^{syn}_i &= \alpha * I^{fg}_{avil} + (1 - \alpha) * I^{bg}_i
\label{eq:paste}
\end{align}
where $\alpha = 1.0$ with foreground and $\alpha = 0.0$ with background.
Finally, we generate synthetic image $I^{syn}_i$ with the pseudo label, as shown in the top row of Fig. \ref{fig:example_PL}.

\subsection{Pos-Data-Augmentation}
We illustrate our \textbf{P}os-\textbf{D}ata-\textbf{A}ugmentation (\textbf{PDA}) method in Algorithm 2, which also contains three steps.

\begin{algorithm}[!h]
\caption{Pos-Data-Augmentation}
\label{alg:Pos-Pseudo-labeling}
\begin{algorithmic}[1]
\Require pedestrian masks $M^{syn}_{j},j=1,2,\cdots,K$, background images $\hat{I}^{bg}_{i},i=1,2,\cdots,N$;
\Ensure $\hat{I}^{syn}_i, i=1,2,\cdots,N$.
\State Fix random seed;
\State Shuffle $\hat{I}^{bg}$;
\State $i \gets 1$;
\While{$i <= N$}
    \State $M^{syn}_{j} \gets$ randomly select from $M^{syn}$ ;
    \State Randomly location $P^{\prime}_{off}$ from freespace of $\hat{I}^{bg}_i$;
    \State $\hat{I}^{syn}_i = \alpha * M^{syn}_{j} + (1 - \alpha) * \hat{I}^{bg}_i$ in $P^{\prime}_{off}$;
    \State Create pseudo label of $I^{syn}_i$;
\State $i \gets i + 1$;
\EndWhile
\State \Return $\hat{I}^{syn}$.
\end{algorithmic}
\end{algorithm}

\subsubsection{Collecting pedestrian masks and background images}
Different from \textbf{ODA}, \textbf{PDA} only requires one source pedestrian image $I_s$ and background images $I^{bg}_{i},i=1,2,\cdots,N$. 
The source image requires pedestrians with a complete body structure and black background, with the assistance of the Labelme tool \cite{russell2008labelme}. 
Moreover, Liquid Warping GAN (AttLWB) \cite{lwb2019} could create different human postures with a reference video, which warps the visible textures of source images to the desired poses.
With the help of AttLWB \cite{lwb2019}, we obtain a series of synthetic pedestrian masks $M^{p}_{j},j=1,2,\cdots, K$ with different pedestrian postures, as shown in Fig. \ref{fig:overview}(c).
Furthermore, we also take the morphological operations OPEN and ERODE to process the synthetic pedestrian masks to remove the noise near the mask's contours. 

\subsubsection{Localization of synthetic posture pedestrians.}
Since posture pedestrians must lie within the freespace region of parking scenes, we first detect the freespace region.
We train a simple semantic segmentation model for freespace region detection and randomly pick one location within the model's freespace prediction.
It is notable that this approach slightly requires time and computational resources, but we apply the procedure to ensure the quality of pseudo labels.

\subsubsection{Local fusion of pedestrian masks}
Finally, we resize the masks at a limited scale and paste the pedestrian masks into the background at the selected freespace location:
\begin{align}
\hat{I}^{syn}_i = \alpha * M^{syn}_{j} + (1 - \alpha) * \hat{I}^{bg}_i.
\label{eq:paste2}
\end{align}

Then, we obtain synthetic posture pedestrians with pseudo labels, as shown in Fig. \ref{fig:overview} (e).  
The bottom row of Fig. \ref{fig:example_PL} illustrates the \textbf{PDA}'s examples.



\section{Experiments}

In this section, we report the baseline results and the results of two proposed data-augmentation techniques on our \textbf{PPD} dataset.
Then, we discuss \textbf{PPD}'s generalization across datasets.
Furthermore, we analyze the effects of \textbf{ODA} and \textbf{PDA} by ablation studies and comparisons.

\subsection{Experimental Settings}
\subsubsection{Implementation Details} 
We conduct experiments with the Pytorch framework on Ubuntu system and employ eight NVIDIA RTX A6000s. The learning rate is set to 0.12, while the momentum and learning decay rates are set to 0.9 and 0.01, respectively. For training, we adopt the stochastic gradient descent (SGD) solver, 48 epochs, and 16 batch size. For our data augmentation experiments, we mix 250,000 augmentation images with the original \textbf{PPD} dataset.
\subsubsection{Evaluation Metrics}
The detection task ought to chase superb targets for the location to ensure pedestrians' safety.
Therefore, we select a high IoU criterion of 0.75 for object detection metrics: Average Precision (AP) and Average Recall (AR).
The high threshold forms a stricter examination to filter more robust models for the pedestrian detection task.

\subsubsection{Baseline methods}
Our baseline detectors contain CenterNet \cite{duan2019centernet} with backbone DLA34 \cite{yu2018deep}, YOLOF \cite{article5}, Faster R-CNN \cite{article7}, Cascade RCNN \cite{article8} and RetinaNet \cite{article3} with ResNet-50 \cite{szegedy2017inception}  backbone. 
All baselines have occupied the field of object detection in recent years. 
To ensure comparability, all baselines utilize the same experimental settings as their release.

\subsubsection{Datasets}
We also choose several public datasets for cross-dataset evaluation: COCO \cite{lin2014microsoft}, KITTI \cite{article22}, CityPersons \cite{article24}, and WiderPerson \cite{article23}, where the last two datasets aim for the category ``Person".

\subsection{Results and Analysis}
\subsubsection{\textbf{Results on PPD and PPD w/ DA}}
To demonstrate the effectiveness of our data-augmentation methods, we conduct baseline evaluation based on the \textbf{PPD} dataset and the mixed dataset with augmentation images, as shown in Table \ref{tab:baselines_result}.
For the original \textbf{PPD} dataset, the two-stage Faster RCNN wins on AP75, and the one-stage CenterNet wins on AR75.
As a new anchor-free pipeline, CenterNet focuses on object center detection, which brings higher recall and perhaps lower precision.
All the performance enhancements are approximately 2\% to 4\% when mixed with data-augmentation images. 
We attribute the advancement to our realistic synthetic images, which satisfy the appetite of pedestrians with various occlusions and postures at a low cost. 

Besides, we explore the subdatasets' performance with data-augmentation images (w/ DA), as shown in Table \ref{tabel:subdataset}. Additional data-augmentation images take effect individually on both sub-datasets. 


\begin{table}[!t]
\caption{The performance of baselines on our \textbf{PPD} dataset without or with data-augmentation images (PPD w/ DA). }
\label{tab:baselines_result}
\centering
\resizebox{\linewidth}{!}{
\begin{tabular}{c|cccccc}
\toprule
\multirow{2}{*}{Method} & \multicolumn{2}{c}{PPD} & \multicolumn{2}{c}{PPD w/ DA} & \multicolumn{2}{c}{Improvement} \\
                        & AP75     & AR75     & AP75               & AR75              & AP75           & AR75       \\ \midrule
CenterNet \cmmnt{\cite{duan2019centernet}}  & 43.4     & 52.7         & 46.6               & 55.7                  & \textbf{+3.2}  & \textbf{+3.0}  \\
YOLOF \cmmnt{\cite{article5}}                   & 52.1     & 40.3         & 54.3              & 43.1                  & \textbf{+2.2}  & \textbf{+2.8}  \\
Faster RCNN \cmmnt{\cite{article7}}            & 54.1     & 43.3         & 57.4               & 45.3                  & \textbf{+3.3}  & \textbf{+2.0}     \\
Cascade RCNN \cmmnt{\cite{article8}}           & 54.0    & 42.7         & 56.9               & 45.1                 & \textbf{+2.9}  & \textbf{+2.4}  \\
RetinaNet \cmmnt{\cite{article3}}               & 51.0     & 41.3         & 54.1               & 44.5                  & \textbf{+3.1}     & \textbf{+3.2}  \\ \bottomrule
\end{tabular}
}
\end{table}


\begin{table}[!t]
\caption{The performance of CenterNet  on sub-datasets without or with data-augmentation images (w/ DA).}
\label{tabel:subdataset}
\centering
\begin{tabular}{l|cccccc}
\toprule
\quad \quad \quad Subdataset
& AP75 & AR75  \\
\midrule
Occlusion Pedestrians  &41.3  & 51.2  \\ 
Occlusion Pedestrians w/ DA &\textbf{44.3}  &\textbf{54.4}     \\
\midrule
Posture Pedestrians  &43.5  & 53.9  \\ 
Posture Pedestrians w/ DA &\textbf{45.7}  &\textbf{56.3}     \\
\bottomrule
\end{tabular}
\end{table}




\begin{table}[!t]
\small
\caption{Evaluation of the generalizability of public datasets to the PPD dataset based on the CenterNet \cite{duan2019centernet} method. 
A$\rightarrow$B means adopting the model pretrained on dataset A to evaluate or finetune on dataset B.}
\label{tab:city_wider_result}
\centering
\begin{tabular}{l|cccccc}
\toprule
\multirow{2}{*}{\quad \quad \quad Dataset } & \multicolumn{2}{c}{Evaluation} & \multicolumn{2}{c}{Finetune}  \\
& AP75     & AR75     & AP75 & AR75 \\

 \midrule
 PPD (from scratch) &\textbf{43.4}  &\textbf{52.7} &- & - \\
 \midrule
 COCO \cmmnt{\cite{lin2014microsoft}}$\rightarrow$PPD &30.9  &34.7  &\textbf{43.3} &\textbf{52.5}\\ 
 KITTI \cmmnt{\cite{article22}}$\rightarrow$PPD &29.3 &33.2  &\textbf{42.5}  &\textbf{52.3}\\ 
 Citypersons \cmmnt{\cite{article24}}$\rightarrow$PPD&31.6 &35.3 &\textbf{43.7}  &\textbf{53.1} \\ 
 WiderPerson \cmmnt{\cite{article23}}$\rightarrow$PPD &32.1 &36.9 &\textbf{43.6}  &\textbf{53.0}  \\ 
\bottomrule
\end{tabular}
\vspace{-3mm}
\end{table}

\subsubsection{\textbf{Cross-dataset Evaluation}}

First, we test how well models, which perform well on commonly used datasets, perform on our \textbf{PPD} dataset.
We train CenterNet models on the public datasets COCO, KITTI, CityPersons, and WiderPerson. 
Then, we infer and evaluate them on the \textbf{PPD} dataset, as shown in Table \ref{tab:city_wider_result}. 
We observe that models pre-trained on public datasets perform suboptimally, indicating their inadequacy with irregular pedestrian fisheye instances.
Furthermore, we finetune these models on the \textbf{PPD} dataset, also as shown in Table \ref{tab:city_wider_result}.  
In comparison to the \textbf{PPD} model trained from scratch, the pre-trained cross-dataset models do not make much advancement, even with a small performance drop similar to the KITTI dataset.
We conjecture that public datasets are insufficient to compensate for the absence of different pedestrians with occlusion and varied postures.
\begin{table}[!t]
\small
\caption{Evaluation of the generalizability of the PPD dataset to public datasets based on CenterNet \cite{duan2019centernet}.
A$\Rightarrow$B means adopting the model pre-trained on dataset A to finetune on dataset B.
}
\label{tab:ablation_2}
\centering
\begin{tabular}{l|cccccc}
\toprule
\quad \quad \quad Dataset
 & AP75 & AR75  \\
\midrule
 COCO \cmmnt{\cite{lin2014microsoft}} &44.7  &38.7  \\ 
 PPD$\Rightarrow$COCO \cmmnt{\cite{lin2014microsoft}} &\textbf{47.2}  &\textbf{42.5}     \\
\midrule
 KITTI \cmmnt{\cite{article22}}&52.3 &56.2  \\ 
 PPD$\Rightarrow$KITTI \cmmnt{\cite{article22}}&\textbf{55.8} &\textbf{59.8}  \\ 
\midrule
 Citypersons \cmmnt{\cite{article24}}&46.5 &57.3 \\ 
 PPD$\Rightarrow$Citypersons \cmmnt{\cite{article24}}&\textbf{49.1} &\textbf{58.8} \\ 
\midrule
 WiderPerson \cmmnt{\cite{article23}}&53.3 &47.9  \\ 
 PPD$\Rightarrow$WiderPerson \cmmnt{\cite{article23}}&\textbf{56.7} &\textbf{53.1}  \\ 
\bottomrule
\end{tabular}
\vspace{-4mm}
\end{table}


Moreover, we conduct generalization experiments from \textbf{PPD} on public datasets based on CenterNet \cite{duan2019centernet}, as shown in Table \ref{tab:ablation_2}.
After finetuning, our \textbf{PPD} dataset gains approximately 2\% to 5\% enhancement compared to the baselines trained on public datasets, especially for AR 75.
\textbf{PPD}'s pedestrian cases cover the usual pedestrian scenes, which considerably increases the recall and lift generalization ability.


\subsubsection{\textbf{Ablation Study}}
We conduct ablation studies for the \textbf{O}cc-\textbf{D}ata-\textbf{A}ugmentation (\textbf{ODA}) and \textbf{P}os-\textbf{D}ata-\textbf{A}ugmentation (\textbf{PDA}) methods based on CenterNet \cite{duan2019centernet}, as shown in Table \ref{tab:ablation_methods}.
(b) and (c) rows show that training only with synthetic images does not make sense because of the data domain shift. 
From rows (d) and (e), our \textbf{ODA} and \textbf{PDA} obviously make great progress, especially \textbf{ODA}, contributing approximately 2\% AP75 improvement.
Both techniques are effective, and the result performs best in combination, as shown in the (f) row.



\begin{table}[t]
\small
\centering
\caption{Ablation study of two data-augmentation techniques, ODA and PDA, based on the CenterNet method.}
\begin{tabular}{c|ccc|ccc}
    \toprule
    &$\text{PPD}$ & $\text{ODA}$ & $\text{PDA}$  & $\text{AP75}$  & $\text{AR75}$ \\
    \midrule
    (a)&\Checkmark    &                &                 & 43.4 & 52.7  \\
    (b)&              & \Checkmark     &                 & 30.6 & 36.7   \\
    (c)&              &                & \Checkmark      & 28.4 & 33.2   \\
    (d)&\Checkmark    & \Checkmark     &                 & 45.6 & 54.3 \\
    (e)&\Checkmark    &                & \Checkmark      & 44.5 & 53.4   \\
    (f)&\Checkmark    & \Checkmark     & \Checkmark      & \textbf{46.6} & \textbf{55.7}   \\
  \bottomrule
\end{tabular}
\label{tab:ablation_methods}  
\end{table}

\begin{table}[!t]
\small
\centering
\caption{Comparison between our data-augmentation method and the copy-paste method \cmmnt{\cite{article50}} with the same CenterNet method, based on the PPD dataset.}
\begin{tabular}{c|c|cc}
    \toprule
	 & Method & $\text{AP75}$ & $\text{AR75}$  \\
    \midrule
    (a) & PPD                           & 43.4 & 52.7    \\
    (b) & PPD w/ Copy-paste \cmmnt{\cite{article50}}   & 37.2 & 48.1 \\
    (c) & PPD w/ DA                     & \textbf{46.6} & \textbf{55.7} \\
   \bottomrule
\end{tabular}
\label{tab:methods}  
\end{table}

\subsubsection{\textbf{Comparison with Copy-paste}}
Copy-paste \cite{article50} plays an important role in hybrid image augmentation.
We compare our data-augmentation techniques with copy-paste based on CenterNet, as shown in Table \ref{tab:methods}.
Surprisingly, from (b) row, the model's performance with copy-paste degrades by approximately  5\%.
We analyze copy-paste copies and paste instances across images but without any fusion processing, which easily leads to unreliable images and false detection. 
In contrast, our well-organized adaptive pasting localization and fusion strategies bring an improvement of 3.2\% in AP75 and 3.0\% in AR75 in the (c) row.

\begin{table}[!t]
\centering
\caption{Exploration of the optimal quantity for data-augmentation approaches using the CenterNet \cmmnt{\cite{duan2019centernet}} method and the PPD dataset. The greatest amount of training data may not always provide the best results.}
\begin{tabular}{c|l|cc}
    \toprule
     & Amount  & $\text{AP75}$ & $\text{AR75}$ \\
    \midrule
    (a)&$+$ 0   &  43.4 & 52.7  \\
    (b) &$+$ 50,000  &  44.1 & 52.9 \\
    (c) &$+$ 150,000 &  45.3 & 53.4 \\
    (d) &\textbf{$+$ 250,000} &  \textbf{46.6} & \textbf{55.7} \\
    (e) &$+$ 350,000 &  46.1 & 54.2 \\
    (f) &$+$ 450,000 &  45.9 & 53.3 \\
    \bottomrule
\end{tabular}
\label{tab:opl_numbers}  
\end{table}

\subsubsection{\textbf{Discussion with amount of data-augmentation images}}
\label{sec:nPseudo}

Theoretically, we could produce infinite pseudo-labeling images. 
However, a large quantity of training data occupies large amounts of resources and time. 
To trade off the efficiency and performance, we explore the optimal quantity for pseudo-labeling images based on the CenterNet \cite{duan2019centernet} method and the \textbf{PPD} dataset, as shown in Table \ref{tab:opl_numbers}. Interestingly, from rows (d), (e) and (f), most training data do not perform the best result, perhaps resulting from overfitting.
Before overfitting, a larger image volume means greater enhancement, as illustrated in rows (b), (c) and (d).
\section{Conclusion}
In this paper, we have presented a new dataset, the Parking Pedestrians Dataset (\textbf{PPD}), as well as two unique data-augmentation techniques, Occ-Data-Augmentation (\textbf{ODA}) and Pos-Data-Augmentation (\textbf{PDA}). By providing a diversity of pedestrian postures, the proposed dataset aims to assist the industry in constructing a more secure advanced driving assistance system. Moreover, we provide two techniques for enhancing pedestrian detection performance using data augmentation. Extensive experiments on the proposed \textbf{PPD}  validate the effectiveness of the techniques. However, \textbf{PPD}  has a large capacity for development, including how to strengthen the realism of the data augmentation, simplify our methodologies, deal with sustainably increasing data and have the potential for diverse vision tasks. Nevertheless, we expect \textbf{PPD}  to inspire more relevant research and promote the performance of pedestrian detection under parking scenes. In the future, the proposed \textbf{PPD} dataset's potential not only lies in pedestrian detection but can also be extended into other vision tasks, such as pixel-wise semantic segmentation, video object detection and 3D object detection tasks.

\bibliographystyle{IEEEtran}
\bibliography{sample-base}

\end{document}